\PassOptionsToPackage{table,dvipsnames}{xcolor}

\documentclass[sigconf]{acmart}

\pdfoutput=1

\usepackage{soul}
\usepackage{url}
\usepackage[utf8]{inputenc}
\usepackage{graphicx}
\usepackage{amsmath}
\usepackage{booktabs}
\usepackage[]{natbib}

\makeatletter%
\@ifclassloaded{acmart}{
  \makeatletter
  \def\mdseries@tt{m}
  \makeatother
}{}
\makeatother

\usepackage{epsfig}
\usepackage{pgfplotstable}
\usepackage{pgfplots}
\usepgfplotslibrary{groupplots}
\usepgfplotslibrary{statistics}
\pgfplotsset{compat=newest} 
\usepackage{amsmath,mathtools}
\usepackage{amssymb}
\usepackage{bbm}
\usepackage{microtype}
\usepackage{tikz}
\usetikzlibrary{decorations.text,calc,shapes,arrows,arrows.meta, positioning,shapes.misc,decorations.markings,decorations.markings,decorations.pathreplacing,matrix,patterns,chains,shapes.geometric,fit}

\usepackage{booktabs}
\usepackage{multirow}
\usepackage{adjustbox}
\usepackage{verbatim}
\usepackage[T1]{fontenc} 
\usepackage[title]{appendix}
\usepackage{adjustbox} 
\usepackage{graphicx}
\usepackage{caption}
\usepackage{subcaption}
\usepackage[ruled,linesnumbered]{algorithm2e}

\usepackage{siunitx} 
\usepackage{etoolbox} 
\robustify\bfseries
\usepackage{nicefrac}
\usepackage[colorinlistoftodos]{todonotes}

\newcommand{\textcite}[1]{\citet{#1}}

\renewcommand{\cite}[1]{\citep{#1}}

\usepackage[autostyle, english=american]{csquotes}
\MakeOuterQuote{"}

\makeatletter
\pgfplotsset{
    every axis x label/.append style={
        alias=current axis xlabel
    },
    legend pos/outer south/.style={
        /pgfplots/legend style={
            at={%
                (%
                \@ifundefined{pgf@sh@ns@current axis xlabel}%
                {xticklabel cs:0.5}%
                {current axis xlabel.south}%
                )%
            },
            anchor=north
        }
    }
}
\makeatother

\newcolumntype{t}{>{\ttfamily}l}
\newcolumntype{T}{>{\ttfamily}c}

\newcolumntype{$}{>{\global\let\currentrowstyle\relax}}
\newcolumntype{^}{>{\currentrowstyle}}

\DeclareMathOperator*{\argmin}{argmin}

\everypar{\looseness=-1} 
\begin{document}


\title{Heterogeneous Relational Kernel Learning}


  \author{Andre T. Nguyen}
  \email{ Nguyen_Andre@bah.com}
  \affiliation{%
    \institution{Booz Allen Hamilton}
    \institution{University of Maryland, Baltimore County}
  }

  \author{Edward Raff}
 \email{ Raff_Edward@bah.com}
  \affiliation{%
    \institution{Booz Allen Hamilton}
    \institution{University of Maryland, Baltimore County}
  }

\copyrightyear{2019}
\acmYear{2019}
\setcopyright{acmlicensed}
\acmConference[MileTS '19]{MileTS '19: 5th KDD Workshop on Mining and Learning from Time Series}{August 5th, 2019}{Anchorage, Alaska, USA}
\acmBooktitle{MileTS '19: 5th KDD Workshop on Mining and Learning from Time Series, August 5th, 2019, Anchorage, Alaska, USA}

\acmPrice{}
\acmDOI{}
\acmISBN{}



\begin{abstract}
Recent work has developed Bayesian methods for the automatic statistical analysis and description of single time series as well as of homogeneous sets of time series data. 
We extend prior work to create an interpretable kernel embedding for heterogeneous time series. Our method adds practically no computational cost compared to prior results by leveraging previously discarded intermediate results. We show the practical utility of our method by leveraging the learned embeddings for clustering, pattern discovery, and anomaly detection. These applications are beyond the ability of prior relational kernel learning approaches.

\end{abstract}

\settopmatter{printfolios=true} 


\maketitle

\section{Introduction}

In science, when presented with a highly complex system to understand, a good first step is to identify and model order present in a subset of the data to develop hypotheses to investigate further. Finding hidden pockets of order is especially difficult in large datasets, and as a result finding these pockets is a task ripe for machine learning. However, machine learning methods typically attempt to model all of the data or most of the data in the presence of outliers. Modeling the majority of the data is inappropriate for hypothesis generation as we are looking for order in small subsets of the data, which entails that most of the data should be considered as outliers by the models we are trying to fit. Solving the problem of automatic hypothesis generation from large, noisy datasets has to potential to accelerate the pace of discovery in the natural and social sciences by automating the formulation of research questions from order hidden deep in the data \cite{WeinsteinInPreparation}.

\textcite{Duvenaud2013StructureSearch,Lloyd2014AutomaticModels} introduce a method for the automatic statistical analysis of time series using compositional Gaussian process kernel search. A time series is modeled by a Gaussian process model and the goal is to find a descriptive and expressive kernel. This approach is capable of automatically discovering underlying structure in a time series such as change points, trends, local and global behaviors, periodicities, and variations at multiple resolutions. Compositional kernel search builds its explanation of the data starting from simple, interpretable concepts (periodicity, linearity, noise, variance, change...) and combining these concepts iteratively to better model the data. The compositional nature of the approach allows for the automatic description of the discovered data characteristics in human-friendly natural language. For example, the product of squared exponential and periodic kernels can be interpreted as “locally periodic” structure, and the addition of squared exponential and periodic kernels can be interpreted as “periodic with noise.” \textcite{Yunseong2016AutomaticSeries} introduces two extensions to the work of \textcite{Duvenaud2013StructureSearch,Lloyd2014AutomaticModels} which allow for the modeling of multiple time series using compositional kernel search. At a high level, the authors of \textcite{Yunseong2016AutomaticSeries} achieve this by assuming either that all the time series share a same kernel or that they are all modeled by kernels that share a common component that should be interpretable, while allowing the remaining unexplained structure to be modeled in a non-interpretable manner. 

Computational tractability is the primary challenge to extending the techniques from \textcite{Duvenaud2013StructureSearch,Lloyd2014AutomaticModels, Yunseong2016AutomaticSeries} to find structure in subsets of the time series as searching through all the possible structure sharing combinations would result in an explosion in complexity. We propose a computationally simple extension to the techniques of \textcite{Yunseong2016AutomaticSeries} to discover interpretable structure in subsets of time series data. In addition to advancing the automatic statistician, our research introduces a new interpretable kernel embedding for time series with applications that include clustering and anomaly detection based on the structural similarities and differences among time series in a dataset.

\section{Related Work}

\subsection{Gaussian Processes}

The Gaussian process (GP) is the generalization of the Gaussian probability distribution to functions. More specifically, a Gaussian process is a collection of random variables, any finite number of which have a joint Gaussian distribution \cite{Rasmussen2006RasmussenLearning}. A Gaussian process is completely specified by its mean function and covariance function:
$f(x) \sim GP(m(x),k(x,x'))$
where
$m(x) = E[f(x)]$ and 
$k(x,x') = E[(f(x)-m(x))(f(x')-m(x'))]$.

A zero mean function is often used as marginalizing over an unknown mean function can be expressed using a zero mean GP with a modified kernel. The structure of the kernel function determines how the Gaussian process model generalizes.

\subsection{Automatic Bayesian Covariance Discovery}
\textcite{Duvenaud2013StructureSearch} define a language of regression models by specifying a set of base kernels capturing different function properties and a set of composition rules that combine kernels to produce other valid kernels. To fit a time series, a greedy search is performed over the space of regression models, where each kernel specified model's parameters are optimized by conjugate gradient descent and where optimized models are compared using the Bayesian Information Criterion (BIC):

$$BIC(M) = -2 \log{p(D|M)} + |M| \log{n}$$

where $M$ is an optimized model, $|M|$ is the number of kernel parameters, $p(D|M)$ is the marginal likelihood of the data $D$, and $n$ is the number of data points. BIC is chosen as the criterion for evaluating kernels because it balances model fit and model complexity while avoiding an intractable integral over kernel parameters \cite{Rasmussen2001OccamsRazor, Schwarz1978EstimatingModel}.

\textcite{Lloyd2014AutomaticModels} introduce the Automatic Bayesian Covariance Discovery (ABCD) algorithm which uses the language of regression models from \textcite{Duvenaud2013StructureSearch} to automatically generate natural language descriptions of time series.

\subsection{(Semi-)Relational Kernel Learning}

\textcite{Yunseong2016AutomaticSeries} introduce two kernel learning methods that extend ABCD to model shared covariance structures across multiple time series. Relational Kernel Learning (RKL) aims to find a model that explains multiple time series $D = {d_1, d_2, ..., d_J}$ well. Assuming conditional independence of the marginal likelihoods of each time series allows for the simple computation of the the marginal likelihood of the entire dataset:

$$p(D|M) = p(d_1, d_2, ..., d_J|M) = \prod_{j=1}^J p(d_j|M)$$

The presence of exactly identical structure across all the time series in a dataset is rare. To accommodate for variation in individual time series within a dataset, Semi-Relational Kernel Learning (SRKL) relaxes the exactly identical structure constraint of RKL by learning a set of kernels, one for each time series in a dataset. The kernels share a common component that captures structure found across the dataset while retaining individual components. In particular, the set of kernels learned by SRKL can be written as 
$${K_j=K_S+K_{d_j} | d_j \in D, j = 1,2,...,J}$$ 
where $K_S$ is the shared kernel component and the $K_{d_j}$ are the individual kernel components.

The authors of \textcite{Yunseong2016AutomaticSeries} note that while it would be ideal to use an ABCD type search over a space of regression models to learn the individual kernel components, doing so is not practically feasible due to the explosion of the search space. To avoid the complexity issues, the individual kernel components are represented using the spectral mixture kernel \cite{Wilson2013GaussianExtrapolation}. While this allows SRKL to model multiple time series that may have some structural differences, the single shared kernel component makes it still necessary that the time series be somewhat homogeneous in nature. This is problematic when outliers exist in the data or when the data is heterogeneous. 

\section{Heterogeneous Relational Kernel Learning}

In this section, we introduce a computationally feasible procedure for uncovering structure found in subsets of time series but not necessarily all time series in the dataset. These time series could be described as heterogeneous, and so we term our method Heterogeneous Relational Kernel Learning (HRKL). The procedure is simple to implement and can readily be incorporated into RKL and SRKL with little additional computation, enabled by the reuse of intermediary computational outputs from RKL. 

Intuitively, if a subset of time series are structurally similar, kernels that explain one of the members of the subset well should explain the entire subset generally well. Similarly, kernels that explain one of the members of the subset poorly should tend to explain the entire subset poorly. 

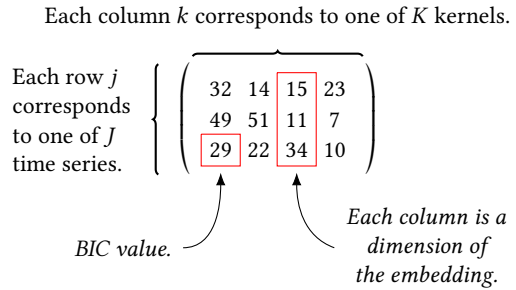
\begin{figure}[!htb]
\centering
\begin{tikzpicture}[
        Brace/.style={
            decorate,
            thick,
            decoration={
                brace,
                amplitude=2pt,
                raise=-7pt
            }
        },]
        \matrix [matrix of math nodes,left delimiter=(,right delimiter=)] (m)
        {
            32 & 14 & 15 & 23 \\               
            49 & 51 & 11 & 7 \\               
            29 & 22 & 34 & 10 \\           
        };  
        \draw[color=red] (m-1-3.north west) -- (m-1-3.north east) -- (m-3-3.south east) -- (m-3-3.south west) -- (m-1-3.north west);
        
        \draw[color=red] (m-3-1.north west) -- (m-3-1.north east) -- (m-3-1.south east) -- (m-3-1.south west) -- (m-3-1.north west);
        
        \draw [LaTeX-] (m-3-3.south) ++(0,-2.5pt) [out=-90,in=160] to ++(5mm,-10mm) node [right, xshift=+0.5mm, font=\itshape, text=black, align=center] {Each column is a\\dimension of\\the embedding.};
        
        \draw [LaTeX-] (m-3-1.south) ++(0,-2.5pt) [out=-90,in=0] to ++(-5mm,-10mm) node [left, xshift=-0.5mm, font=\itshape, text=black, align=center] {BIC value.};
        \node[inner xsep=20pt,inner ysep=0pt,fit=(m)](A){};
        \node[inner xsep=00pt,inner ysep=10pt,fit=(m)](B){};
        \draw[Brace] (A.180 |- A.270) -- (A.180 |- A.90) node[midway,left]{\shortstack[l]{Each row $j$\\corresponds\\to one of $J$\\time series.} };
        \draw[Brace] (B.90 -| B.180) -- (B.90 -| B.0) node[midway,above]{\shortstack[l]{Each column $k$ corresponds to one of $K$ kernels.}};

    \end{tikzpicture}
  \caption{
  Illustration of the HRKL time series embedding.}
  \label{fig:embedding}
\end{figure}

Our extension of RKL is simple. Instead of using BIC values for determining only the best model, we save the BIC value for every kernel-series combination evaluated during the RKL search process. After an iteration of searching over $K$ kernels to fit $J$ time series a $J$ by $K$ BIC history matrix $B$ can be defined. Each matrix element $B_{j,k}$ corresponds to the BIC of a Gaussian process model specified by kernel $k$, optimized for time series $d_j$. This matrix $B$ is illustrated in \autoref{fig:embedding}. Each row of the BIC history matrix is then standardized by removing the mean and scaling to unit variance. This standardized matrix is used as the time series embedding. Each row of the BIC history matrix corresponds to the representation of a time series in the embedded space, and each column is a dimension of the embedded space and is associated with a specific kernel. We note that each dimension of the embedding is interpretable because if the language of regression models from \textcite{Duvenaud2013StructureSearch} is used, then each dimension of the embedding corresponds to an interpretable kernel composed of interpretable base kernels. The above explanation fully describes our new approach. Despite its simplicity, it allows us to extend RKL to easily handle time series of a heterogeneous nature, which we demonstrate with a number of experiments. HRKL is summarized in \autoref{alg:hrkl}.

\begin{algorithm}[]
 \KwIn{Multiple time series $D=d_1, \ldots,d_M$, initial kernel $k$, initial hyperparameters $\theta$, expansion grammar $G$.}
 \KwOut{$\forall i \in [1, C] , (D_i, k_i)$ where $D_i$ is the set of time series in cluster $i$, and $k_i$ is the kernel that best describes cluster $i$.}
 
  \tcp{Use initial kernel and expansion grammar to generate list of kernels to evaluate.}
 $K \leftarrow \text{expand}(k, G)$\;
 
 \tcp{Initialize BIC history matrix.}
  $B \leftarrow \text{array}(M,\text{length}(K))$\;
 
 \tcp{For each kernel.}
 \For{$k \in K$}{
    \tcp{For each time series in the dataset.}
    \For{$d \in D$}{
        \tcp{Fit Gaussian process.}
        $k(\theta) \leftarrow \argmin_{\theta} - \log p(d|k)$\;
        \tcp{Compute and save BIC value.}
        $B_{d,k} = \text{BIC}(d,k)$\;
    }
 }

\tcp{Cluster BIC history matrix using columns as features.}
$\{D_1, \ldots,D_C\} \gets \text{cluster}(B)$\;

\tcp{For each cluster of time series.}
\For{$D_c \in \{D_1, \ldots,D_C\}$}{
    \tcp{Find the kernel that best describes the cluster.}
    $k_c \leftarrow \argmin_{k \in K} \text{BIC}(D_c,k)$\;
}
\KwRet{$[(D_1,k_1),\ldots,(D_C,k_C)]$}
\caption{Heterogeneous Relational Kernel Learning}
\label{alg:hrkl}
\end{algorithm}

\section{Experiments}

We run three experiments to explore the properties and behavior of our interpretable kernel embedding. In particular, we aim to elucidate the strengths and weaknesses of the embedding to understand what applications might benefit from the use of HRKL. 

\subsection{Clustering} 

We begin the evaluation of our interpretable kernel embedding on a clustering task aimed at characterizing what is considered as similar in the embedding space.

\subsubsection{Data}
\begin{figure}[!htb]
\centering
\includegraphics[width=\linewidth]{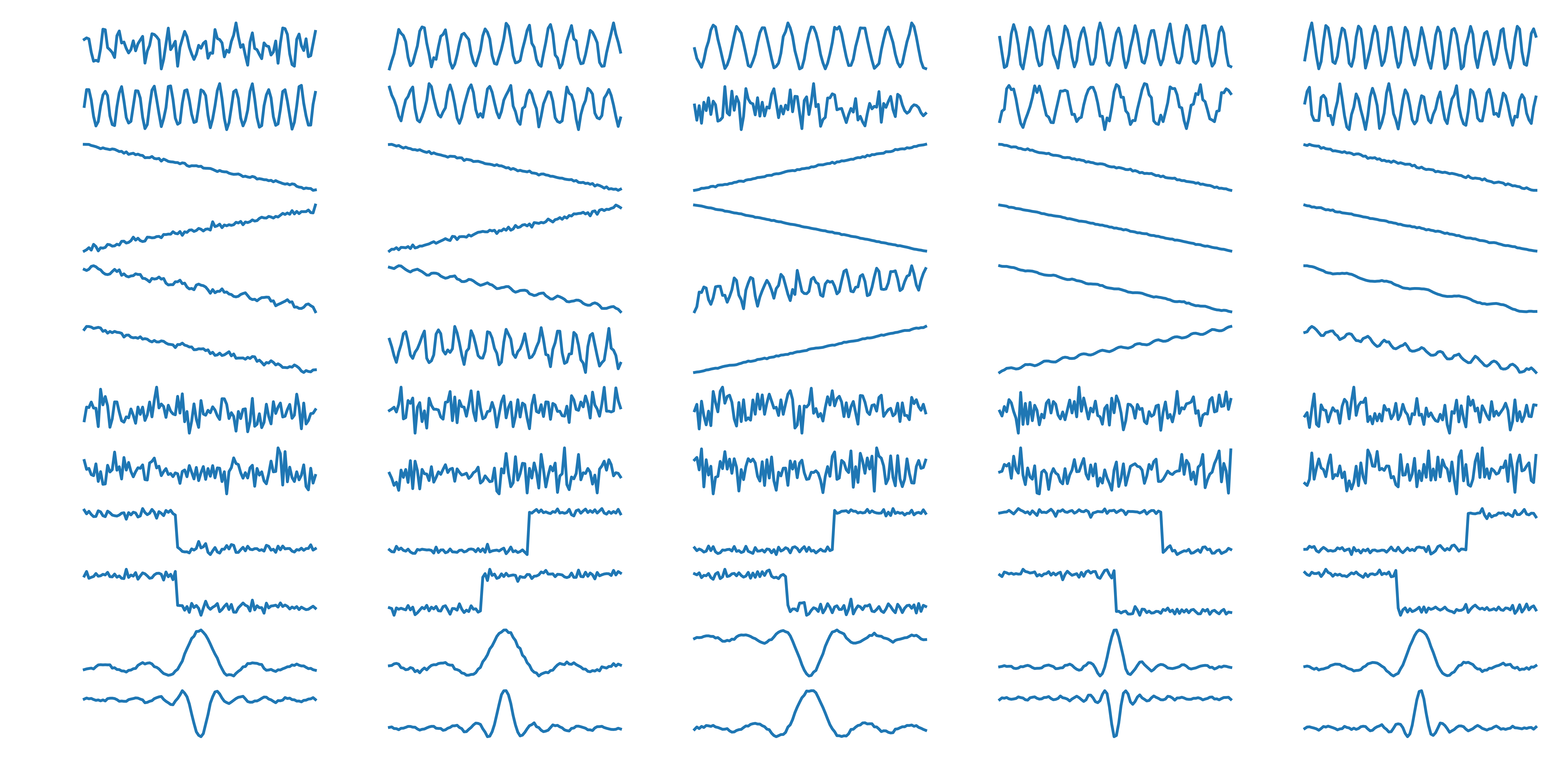}
  \caption{
  Dataset consisting of 60 time series
   with heterogeneous structure that can't be modeled well by (S)RKL. 
}
  \label{fig:data_synthetic}
\end{figure}

We generate a synthetic dataset consisting of 60 standardized time series as shown in \autoref{fig:data_synthetic}. The first 10 series are sine waves with varying amplitudes, frequencies, phases, and noise levels. The next 10 are lines with varying slopes, intercepts, and noise levels. The next 10 are sine waves with linear trends. The next 10 are random noise. The next 10 are variations on the Heaviside step function. The last 10 time series are variations on the sinc function. Each set of 10 time series are considered to form a class. The composition of these classes capture the idea that time series can be considered to be similar if they share structural elements, even if the elements differ in parameter values. These six classes will be used as ground truth labels in the evaluation of the clustering task.

\subsubsection{Clustering Methodology} 
Pairwise distances between the rows of the BIC history matrix, $B$, are computed using cosine distance to obtain a $J$ by $J$ distance matrix $P$. We use this distance matrix $P$ to uncover clusters of time series. Cosine distance is used because vector orientation is much more important than vector magnitude when trying to capture the intuition that if a subset of time series are structurally similar they should be well described by a common subset of kernels and poorly described by another common subset of kernels.

Multiple approaches could be used for clustering. We use HDBSCAN, a density-based, hierarchical clustering algorithm which improves upon DBSCAN \cite{Campello2013Density-BasedEstimates}. We use HDBSCAN because of its high cluster stability and because it does not require the specification of the number of clusters beforehand.

\subsubsection{Kernels} 

We use as base kernels the squared exponential kernel, the linear kernel, and the periodic kernel. We then generate a list of 87 kernels to evaluate by taking all non-redundant kernel structures of the following forms where $k_a$, $k_b$, and $k_c$ are base kernels: $k_a$, $k_a * k_b$, $k_a + k_b$, $(k_a * k_b) * k_c$, $(k_a + k_b) * k_c$, $(k_a * k_b) + k_c$, and $(k_a + k_b) + k_c$.

\subsubsection{Baselines} 

We also evaluate three baseline approaches to highlight the differences between our approach and most previous approaches.

Dynamic Time Warping (DTW) measures similarity between time series by non-linearly warping the series in the time dimension \cite{Salvador2004FastDTWSpace}. We use Euclidean distance DTW with HDBSCAN for clustering. 

Symbolic Aggregate approXimation Bag-of-Patterns (SAX BoP) is a histogram-based representation for time series data which is essentially a bag-of words model of the quantized time series. The SAX BoP representation can then be used to compute a pairwise distance matrix followed by clustering. We use the recommended SAX BoP hyperparameter settings from \textcite{Lin2009FindingRepresentation} with Euclidean distance and HDBSCAN for clustering.

The k-Shape algorithm is a stronger baseline as a time-series clustering algorithm that is invariant to scaling and shifting \cite{Paparrizos2017FastClustering}. k-Shape is centroid-based with a distance measure based on the cross-correlation measure. We note that k-Shape requires that the number of clusters be specified beforehand, a requirement that is not shared by our method nor by the other baselines. Knowledge of the number of clusters ahead of time gives k-Shape an inherent advantage over the other methods in the context of our evaluation. 

\subsubsection{Evaluation Metrics} 
We use homogeneity, completeness, and V-measure as cluster evaluation metrics given the known labels for our six classes \cite{Rosenberg2007V-MeasureMeasure}. The homogeneity score captures how well the clustering reflects the desired property that each cluster contains only members of a single class. The completeness score captures how well the clustering reflects the desired property that all members of a given class are assigned to the same cluster. The V-measure is the harmonic mean of the homogeneity and completeness scores.

\subsubsection{Results and Discussion} 

\begin{table}[]
\caption{Clustering performance metrics comparing HDBSCAN using HRKL, DTW, and SAX BoP, as well as k-Shape. Homogeneity, completeness, and V-measure are all bounded below by 0 and above by 1, where 1 corresponds to a perfect clustering.}
\label{tbl:clus_table}
\begin{tabular}{l|llll}
 & Homogeneity & Completeness  & V-Measure  \\ \hline 
 HRKL & \textbf{0.820} & \textbf{0.852} & \textbf{0.836}  \\
 DTW & 0.496 & 0.627 & 0.553 \\
 SAX BoP & 0.363 & 0.684 & 0.475  \\
 k-Shape & 0.490 & 0.526 & 0.507
\end{tabular}
\end{table}

\autoref{tbl:clus_table} summarizes the homogeneity, completeness, and V-measure metrics of the clustering of the data described in \autoref{fig:data_synthetic} using HRKL with HDBSCAN, DTW with HDBSCAN, SAX BoP with HDBSCAN, and k-Shape. HRKL performs the best by a wide margin on all metrics.

An examination of the cluster assignments made by HDBSCAN with our interpretable kernel embeddings also provides insights into the behavior of the embedding. Five clusters were found. The lines, random noise, Heavyside step variations, and sinc function variations ground truth classes were perfectly clustered. A single error was made in the clustering of members of the sine waves class, where the series shown in second row and middle column of \autoref{fig:data_synthetic} was assigned the same cluster as the members of the random noise class. We believe this is a reasonable mistake to make as the high noise level in that particular series made it look very similar to random noise. The class that our method had difficulty with was the third, sine waves with linear trends, class. The majority of the members of this class were clustered with members of the lines class, followed by members of this class being labeled as outliers or clustered with the sine wave class. Again, we believe that these mistakes are somewhat reasonable as members of the sine waves with linear trends class were clustered with members of the sine waves class and lines class. In contrast to our method, the DTW, SAX BoP, and k-Shape baselines all fail to distinguish sine waves from random noise, consistently clustering members of the sine wave and random noise classes together.

We compared to SAX BoP, DTW, and k-Shape to illustrate how methods aimed at discriminating between time series that share a model class but have different model parameters are less appropriate than HRKL for discriminating between time series that are best described by different model classes.

These results validate our intuition that interpretable kernel embeddings are unique in how they consider time series to be similar if the time series share structural elements.

In the context of an automatic statistician, HRKL improves upon RKL and SRKL in the presence of heterogeneous time series data. In particular, RKL and SRKL are both forced to select a \emph{single kernel} to describe the \emph{entire} dataset, while HRKL is able to select a \emph{separate kernel for each sub-population} resulting in more useful descriptions of the data. 
By iterating through each cluster HRKL finds and looking at the selected kernel, we can interpret the HRKL results with the same ease as RKL. We run through this as an example below to denote how HRKL's interpretation captures the heterogeneous nature of the data. 

When run on the data, RKL and SRKL both select the kernel $\text{PER} * \text{SE} + \text{LIN}$ for the \emph{entire} dataset. This kernel would be described in the language of \textcite{Lloyd2014AutomaticModels} as encoding the following additive components: `a linear function' and `a periodic function whose shape changes smoothly'. While this kernel describes the sine waves with linear trends sub-population well, it is not an appropriate description for the majority of the dataset. 

The HRKL sub-population discovery and kernel selection procedure leads to the selection of the following kernels and descriptions for each of the five sub-populations found. 

For the sub-population containing mostly sine waves, the kernel $\text{PER} * \text{PER} + \text{SE} * \text{PER}$ is selected, encoding the additive components: `a periodic function modulated by a periodic function' and `a periodic function whose shape changes smoothly'. The periodic nature of sine waves is well captured by the selected kernel. For the sub-population containing random noise and one sine wave with high noise, the same kernel, $\text{PER} * \text{PER} + \text{SE} * \text{PER}$, is selected.

For the sub-population containing mostly lines as well as sine waves with linear trends, the kernel $\text{LIN} + \text{PER} * \text{SE}$ is selected, encoding the additive components: `a linear function' and `a periodic function whose shape changes smoothly'. The characteristics of the sub-population, linear trends sometimes with a periodic trend, are well captured by the selected kernel.

For the sub-population containing step functions, the kernel $\text{SE} + \text{PER} * \text{SE}$ is selected, encoding the additive components: `a smooth function' and `a periodic function whose shape changes smoothly'. 

Finally, the sub-population containing sinc function is described by the $\text{PER} + \text{SE}$ kernel which encodes the additive components: `a periodic function' and `a smooth function'.

The use of our interpretable kernel embeddings leads to a more precise and useful automatic description of heterogeneous time series data as it allows for the uncovering and characterization of sub-populations.

\subsubsection{Running Time and Complexity}

HRKL is able to handle sub-population structure discovery at practically the same computational cost as using RKL to find a single kernel shared by all time series. The cost is practically the same because the $O(n \log n)$ complexity of running a clustering algorithm like HDBSCAN once at the end of the search is far smaller than the complexity of fitting a Gaussian process to evaluate a kernel at $O(n^3)$ complexity. 

In terms of running time on a single CPU, the clustering component of HRKL takes on average 3935 microseconds with a standard deviation of 114 microseconds, while a single kernel evaluation takes on average 63984 microseconds with a standard deviation of 9677 microseconds. In other words, HRKL achieved sub-population discovery at the additional cost of 1/16 of the cost of a single kernel evaluation or at the additional cost of 0.07 percent relative to all of the kernel evaluations. 

HRKL improves significantly from a computational complexity perspective in comparison to RKL and SRKL. As Hwang et al note, SRKL is suboptimal from an interpretability perspective because the spectral mixture kernel is used to model variance unexplained by the kernel shared by all time series, but SRKL needs to be used as learning distinctive interpretable kernels for all time series is computationally unfeasible. In particular, the RKL search could be performed to discover one shared kernel and $n$ distinctive kernels, but the search space explodes in terms of complexity as $O(k^{n+1})$ where $k$ is the number of possible kernels on each search grammar tree for every depth. On top of this, using RKL to discover subsets of time series that share common structure results in a combinatorial explosion as each possible subset combination needs to be evaluated using RKL.

\subsection{Pattern Discovery} 

Next, we evaluate our interpretable kernel embedding on a pattern discovery task. We find HRKL and DTW both perform best and discover the same overall structures. 

\subsubsection{Data}

\begin{figure}[!htb]
\centering
\includegraphics[width=\linewidth]{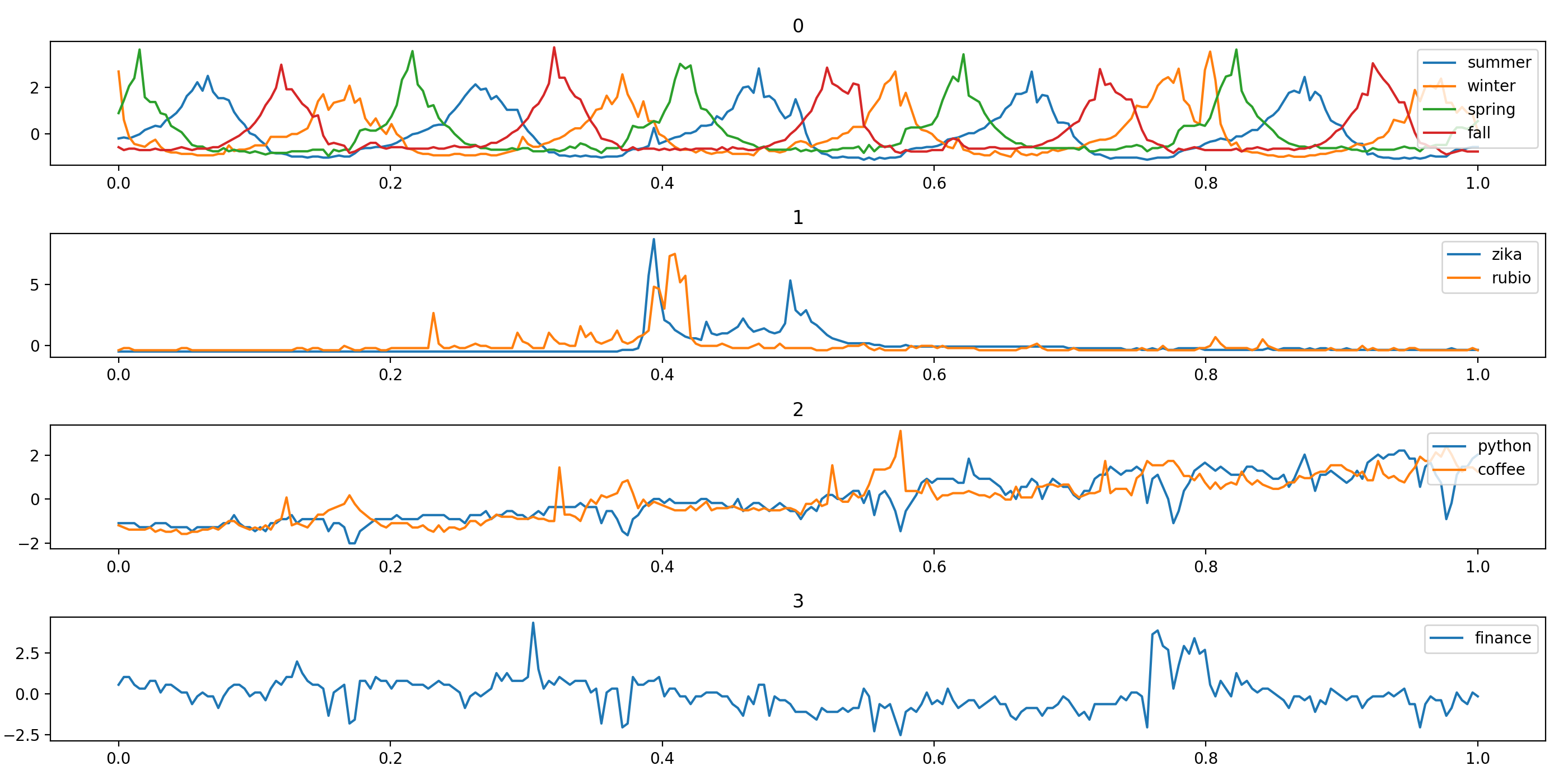}
  \caption{Standardized search volume time series from Google Trends. The search volumes represent relative weekly search popularity in the United States from 2/16/14 to 2/3/19. The data can is divided into four structural subsets.}
  \label{fig:data_gt}
\end{figure}

We use a set of nine search volume time series from Google Trends for the following terms: summer, winter, spring, fall, Zika, Rubio, python, coffee, and finance. The search volumes represent relative weekly search popularity in the United States from 2/16/14 to 2/3/19. The standardized time series are shown in \autoref{fig:data_gt}. The data can be divided into four structural subsets. The search terms representing seasons have a periodic structure, "zika" and "Rubio" are overall flat with temporary surges in interest, "python" and "coffee" have linearly increasing trends, and "finance" has a flat structure with a couple of small surges in interest.   

\subsubsection{Methodology}

We employ a similar methodology to the clustering task in the pattern discovery task, using the same kernels and baselines. As multiple plausible groupings of the data exist and to elucidate what the different approaches consider as similar, we use a hierarchical agglomerative clustering (HAC) algorithm. HAC builds a binary tree over the data by first assigning each datum to its own cluster and then merging groups together. The HAC algorithm maintains an active set of clusters and at each stage determines which two clusters to merge, their union is added to the active set, and they are each removed from the active set. The tree is constructed by keeping track of which clusters are merged together at each step. In order to determine which clusters to merge, the HAC algorithm chooses the pair of clusters in the active set that have the smallest dissimilarity or distance. For the distance metric we choose a single linkage criterion which looks at the euclidean distance between the nearest members of the clusters. A dendrogram can then be used to visualize the computed clustering.
    
\subsubsection{Results and Discussion} 

\begin{figure}[!htb]
    \centering
    
    \begin{subfigure}[b]{0.45\textwidth}
        \centering
        \includegraphics[width=\linewidth,height=140pt]{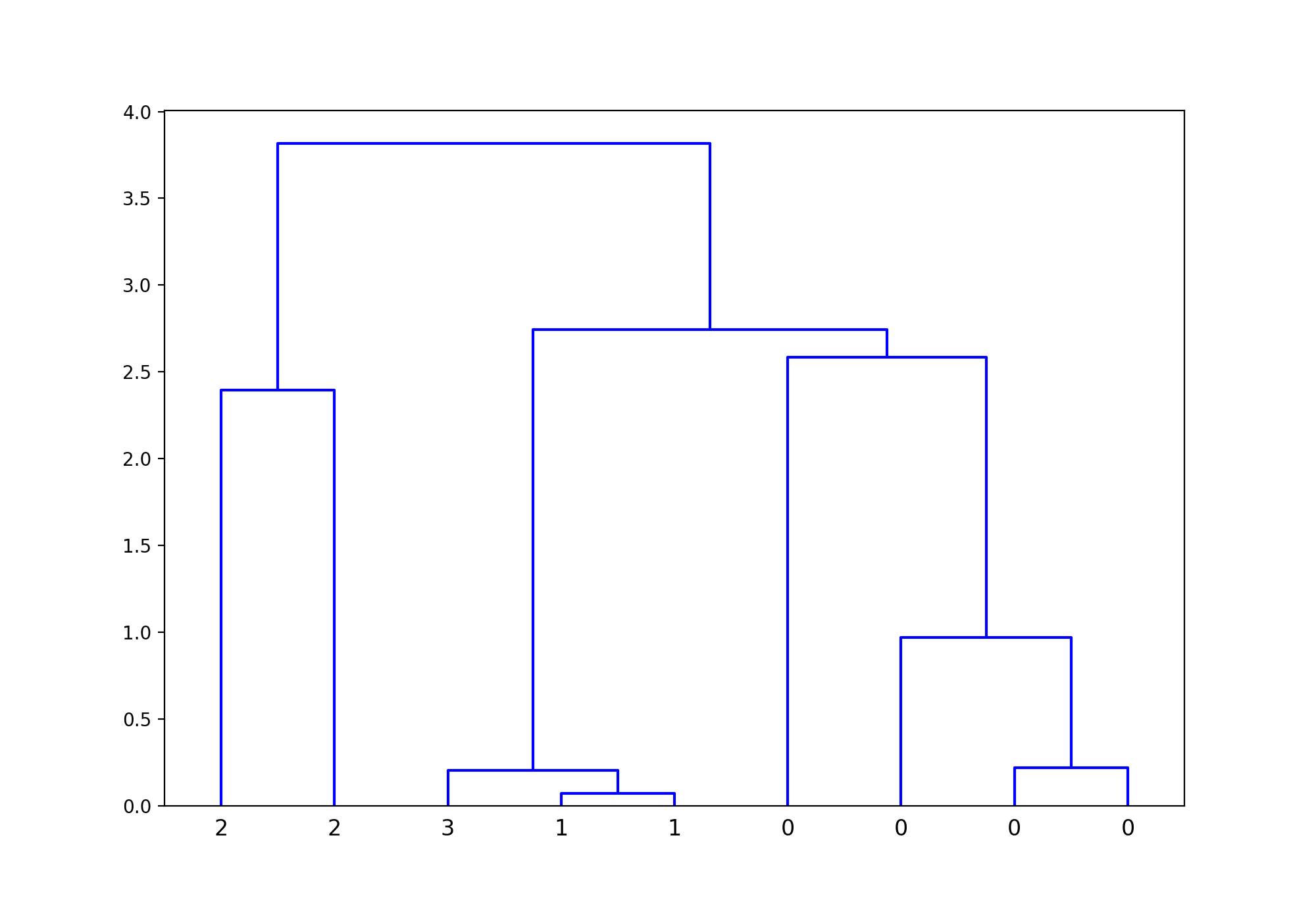}
        \caption{HRKL}
        \label{fig:googletrends-combined-nosrkl-run2_hac_gp}
    \end{subfigure}
    
    \begin{subfigure}[b]{0.45\textwidth}
        \centering
        \includegraphics[width=\linewidth,height=140pt]{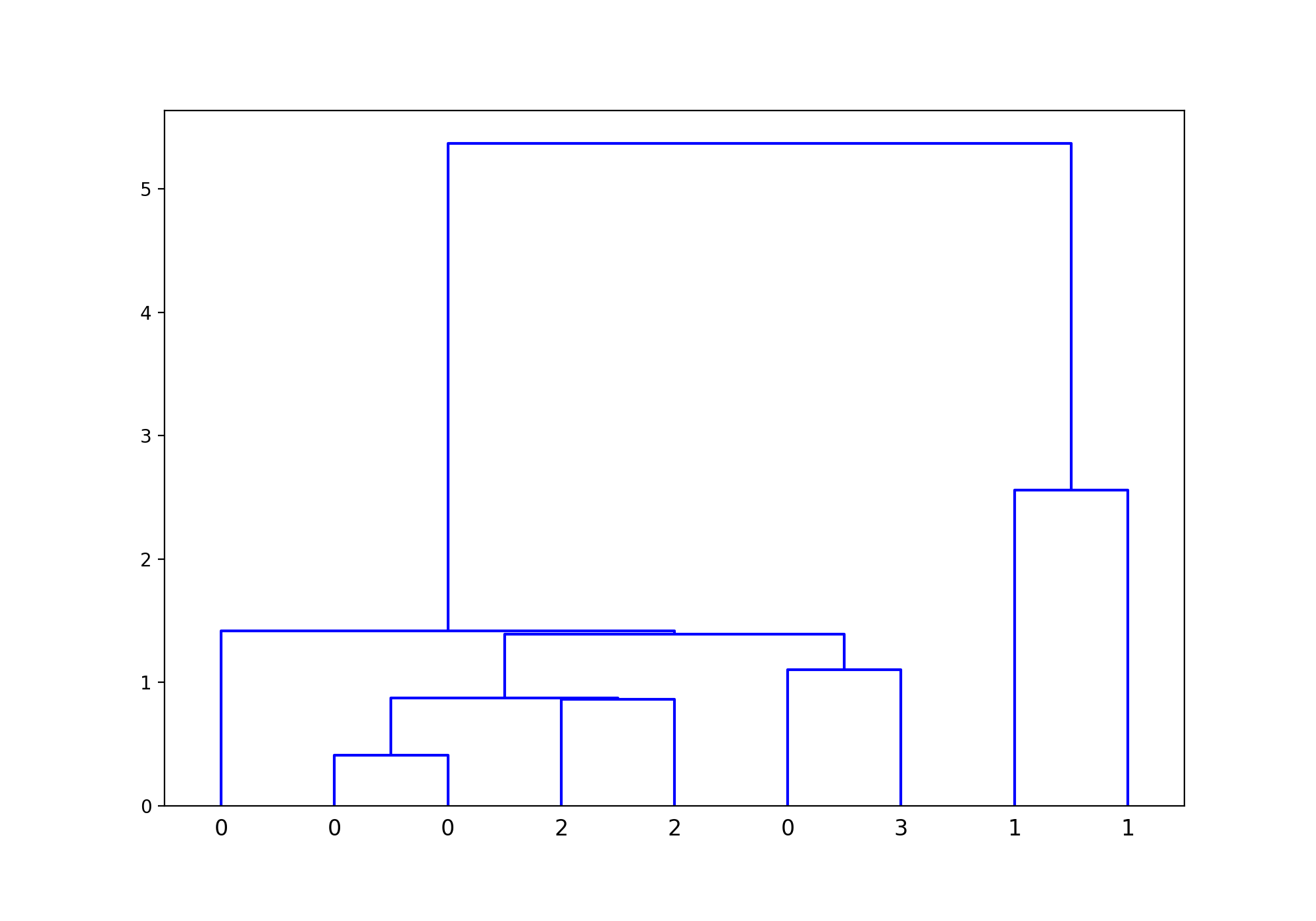}
        \caption{SAX BoP}
        \label{fig:googletrends-combined-nosrkl-run2_hac_sax}
    \end{subfigure}
    
    \begin{subfigure}[b]{0.45\textwidth}
        \centering
        \includegraphics[width=\linewidth,height=140pt]{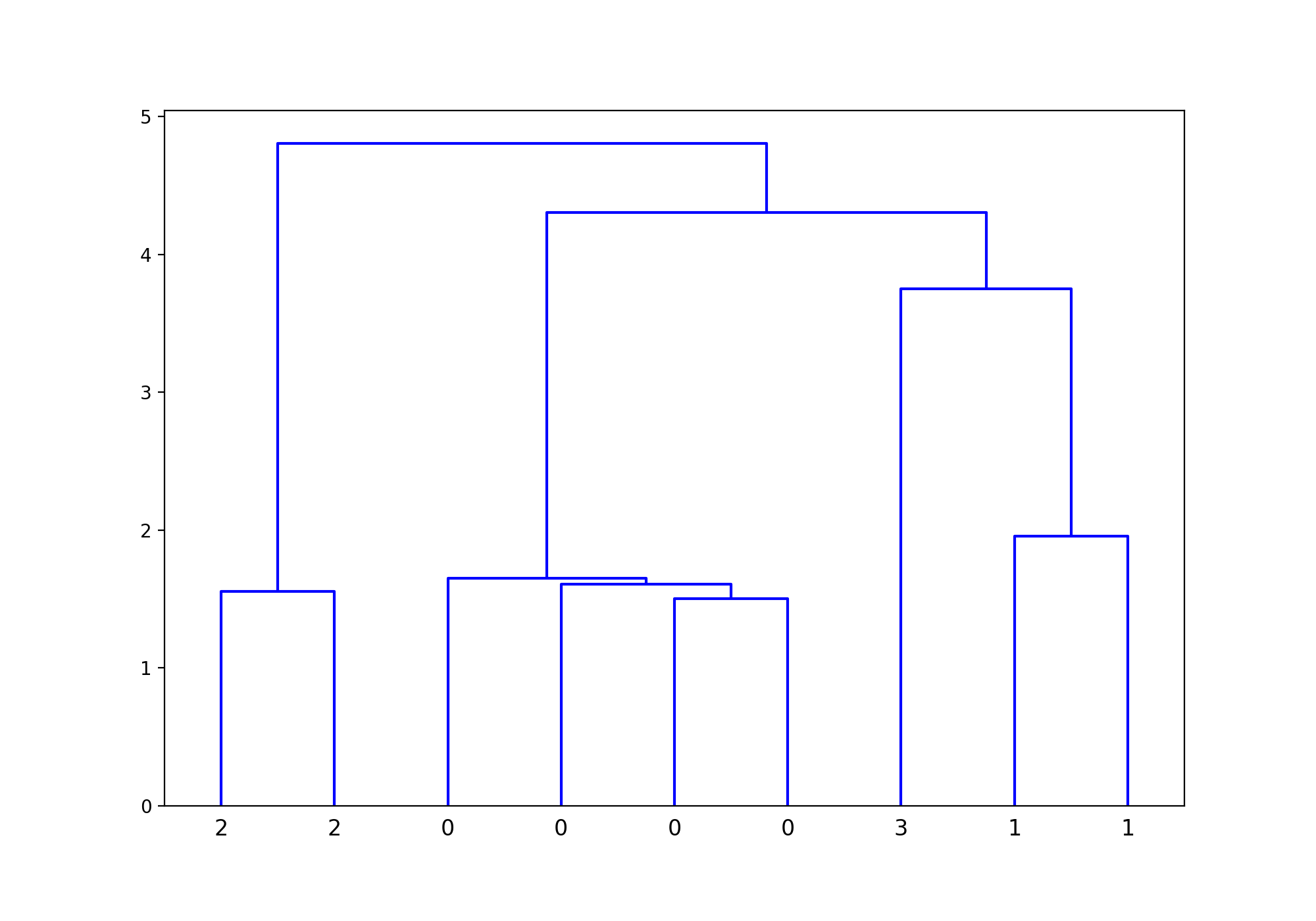}
        \caption{DTW}
        \label{fig:googletrends-combined-nosrkl-run2_hac_dtw}
    \end{subfigure}
    
    \caption{Dendrogram visualizing the HAC clustering of the Google Trends data found using (a) HRKL, (b) SAX BoP, and (c) DTW.  The leaf labels correspond to the grouping labels from \autoref{fig:data_gt}. }
    
\end{figure}

\autoref{fig:googletrends-combined-nosrkl-run2_hac_gp} illustrates the clustering found using our HRKL method, where leaf labels correspond to the grouping labels from \autoref{fig:data_gt}. \autoref{fig:googletrends-combined-nosrkl-run2_hac_sax} shows the clustering found using SAX BoP, and \autoref{fig:googletrends-combined-nosrkl-run2_hac_dtw} shows the clustering found using DTW. As a centroid-based algorithm, k-Shape is not amenable to a dendrogram representation and requires that the number of clusters be specified beforehand. When initialized with the number of clusters set to four, k-Shape recovers the same groupings as our method HRKL, but has the unfair advantage of knowing that there are four clusters.

\autoref{tbl:clus_table_gt} summarizes the clustering performance metrics on the Google Trends data comparing HAC clustering using HRKL, DTW, and SAX BoP, as well as k-Shape, all provided with the correct number of clusters. DTW performs the best, followed by HRKL and k-Shape with a tie. SAX BoP does poorly on this task.

An examination of \autoref{fig:googletrends-combined-nosrkl-run2_hac_gp} shows that the use of HRKL leads to a clustering structure that immediately groups "Zika" and "Rubio", the time series with spikes in search volumes but overall flat structures. These two time series are then grouped with "finance", a time series with an overall flat structure and a number of relatively less significant spikes. The seasons "fall", "winter", "spring", and "summer" are grouped together, and the series with linear trends "python" and "coffee" are also grouped together. Overall on this dataset, the use of our interpretable kernel embedding results in logical groupings of the time series that would allow for heterogeneous relational kernel learning without resulting in an explosion of the search space. 

\autoref{fig:googletrends-combined-nosrkl-run2_hac_sax} shows that while SAX BoP does a good job at grouping "Zika" and "Rubio", SAX BoP is not effective at finding the structure in the rest of the data, for example not being able to uncover the shared periodic structure in the seasonal data. On the other hand, \autoref{fig:googletrends-combined-nosrkl-run2_hac_dtw} shows that DTW led to a nearly identical HAC clustering as the use of our interpretable kernel embedding.

\begin{table}[]
\caption{Clustering performance metrics on the Google Trends data comparing HAC clustering using HRKL, DTW, and SAX BoP, as well as k-Shape, all provided with the correct number of clusters.}
\label{tbl:clus_table_gt}
\begin{tabular}{l|llll}
 & Homogeneity & Completeness  & V-Measure  \\ \hline 
 HRKL & 0.833 & 0.809 & 0.821  \\
 DTW & \textbf{1.000} & \textbf{1.000} & \textbf{1.000} \\
 SAX BoP & 0.470 & 0.597 & 0.526  \\
 k-Shape & 0.833 & 0.809 & 0.821
\end{tabular}
\end{table}

As RKL and SRKL do not result in an embedding but instead a single shared kernel, they cannot be used for the pattern discovery task. This demonstrates the improvement of HRKL over RKL and SRKL, as HRKL can be used for tasks that RKL and SRKL cannot be used for, and HRKL performs better overall than the other methods that can do these tasks.

\subsection{Anomaly Detection} 

We also evaluate our interpretable kernel embedding on an anomaly detection task. In this task, we find HRKL and SAX BoP both solve the task with equal performance, but DTW and k-Shape have significantly degraded results. 

\subsubsection{Data}

\begin{figure}[!htb]
\centering
\includegraphics[width=\linewidth]{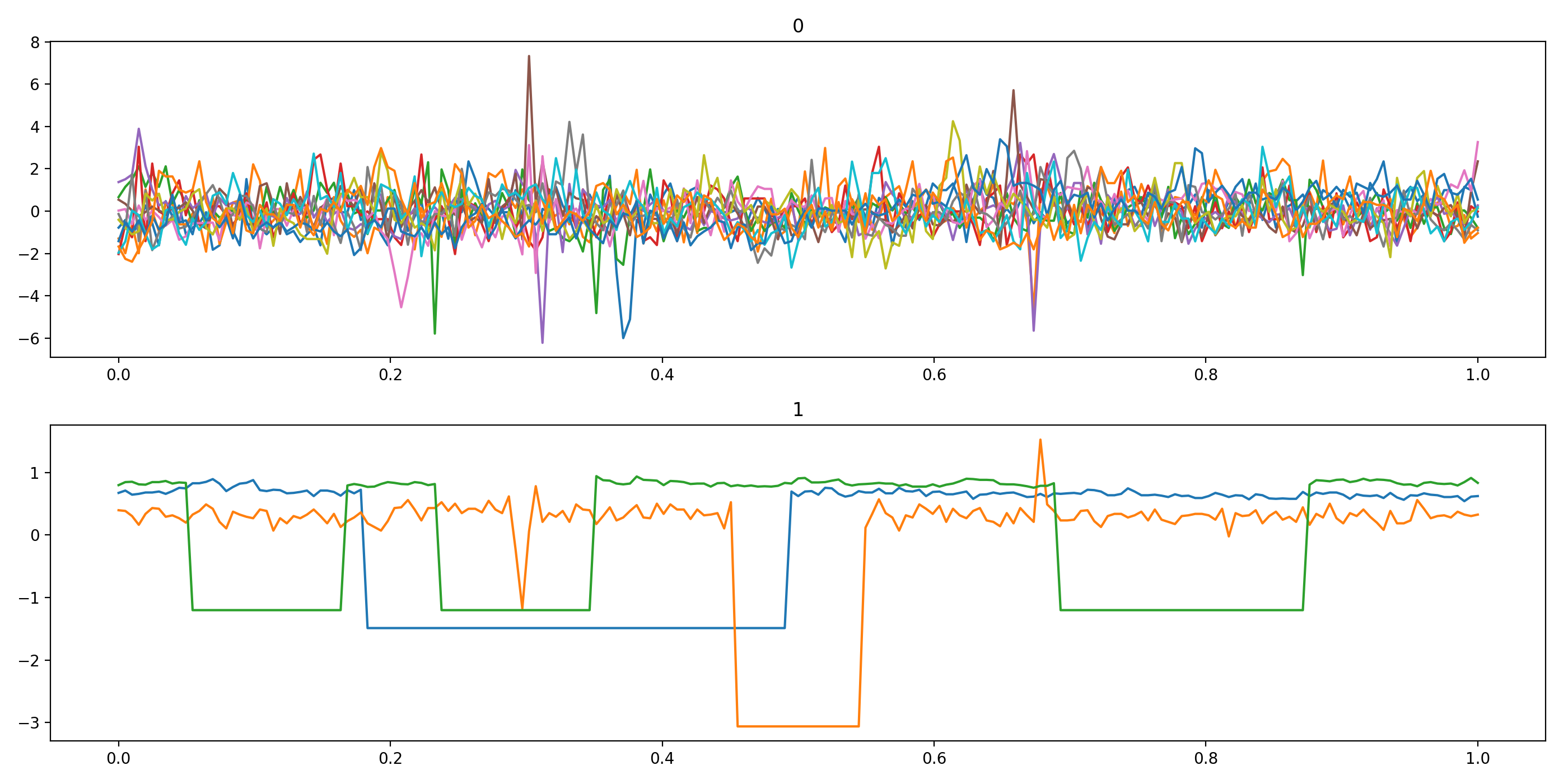}
  \caption{PhysioNet Gait in Aging and Disease dataset which consists of walking stride interval time series for 15 subjects. Three time series are corrupted, where corruption consists of a zeroing out of sections of the series. The uncorrupted time series are shown in the top panel labeled 0, and the corrupted time series are shown in the bottom panel labeled 1.}
  \label{fig:data_corrupted}
\end{figure}

We use the PhysioNet Gait in Aging and Disease dataset which consists of walking stride interval (the time between successive
heel strikes of the same foot) time series for 15 subjects: 5
healthy young adults, 5 healthy old adults, and 5 older adults with Parkinson's
disease. We then randomly select one time series from each class to corrupt, where corruption consists of a zeroing out of sections of the series. This simulates the effect of real world errors that often occur during the reading, processing, transmission, writing, and storage of sensor data \cite{BoozAllenHamilton2013TheScience}. \autoref{fig:data_corrupted} shows both the uncorrupted and corrupted data.

\subsubsection{Methodology}
For our anomaly detection experiment, we use an identical methodology to the one used for the pattern discovery experiment. The goal now is to uncover the corrupted data which should be modeled differently from the uncorrupted data.  

\subsubsection{Results and Discussion} 

\autoref{tbl:clus_table_corrupted} summarizes the clustering performance metrics on the corrupted gait data comparing HAC clustering using HRKL, DTW, and SAX BoP, as well as k-Shape, all provided with the correct number of clusters. HRKL significantly outperforms all of the other methods by a wide margin.  

\begin{figure}[!htb]
    \centering
    
    \begin{subfigure}[b]{0.45\textwidth}
        \centering
        \includegraphics[width=\linewidth,height=140pt]{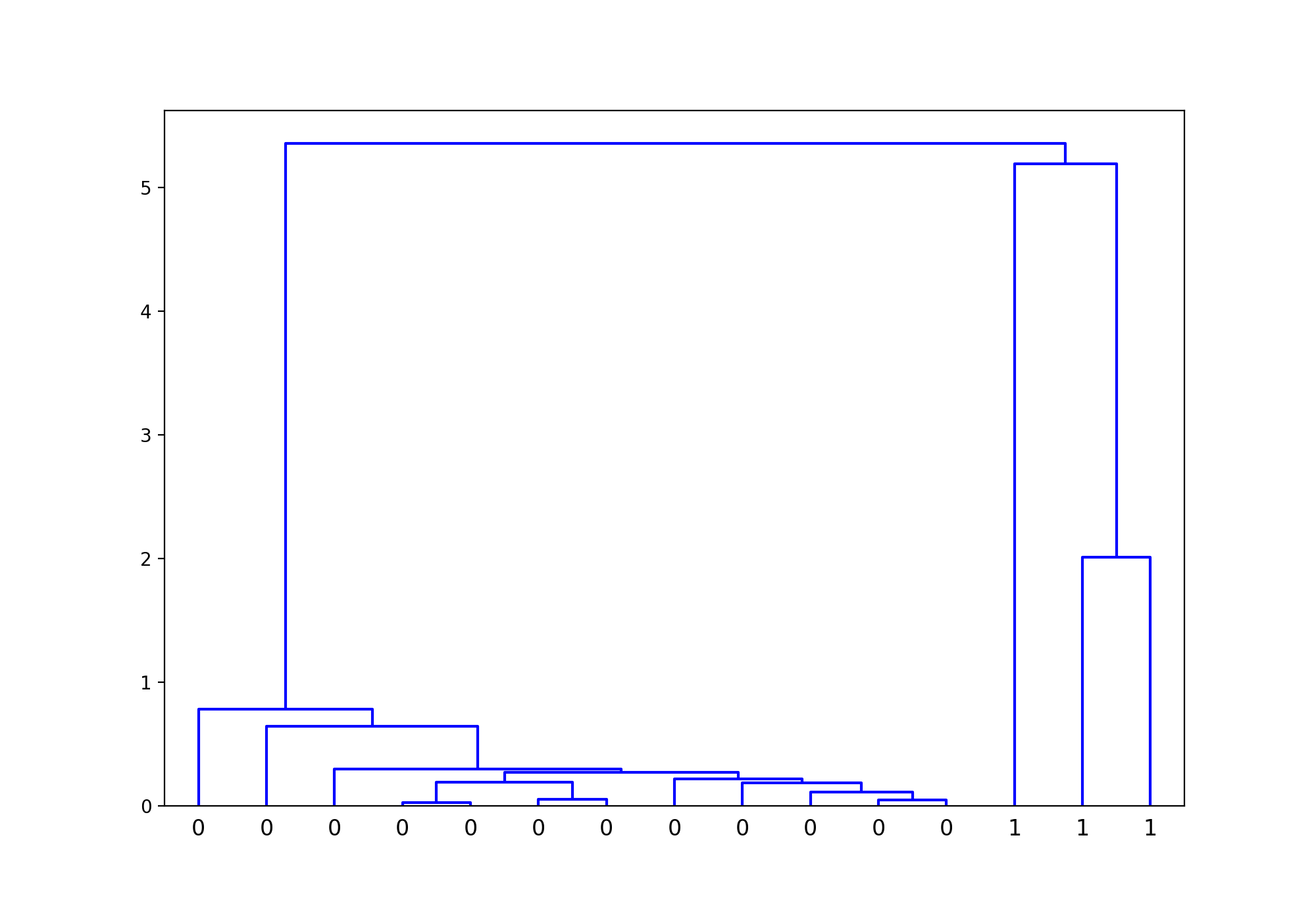}
        \caption{HRKL}
        \label{fig:corrupt-nosrkl-run2_hac_gp}
    \end{subfigure}
    
    \begin{subfigure}[b]{0.45\textwidth}
        \centering
        \includegraphics[width=\linewidth,height=140pt]{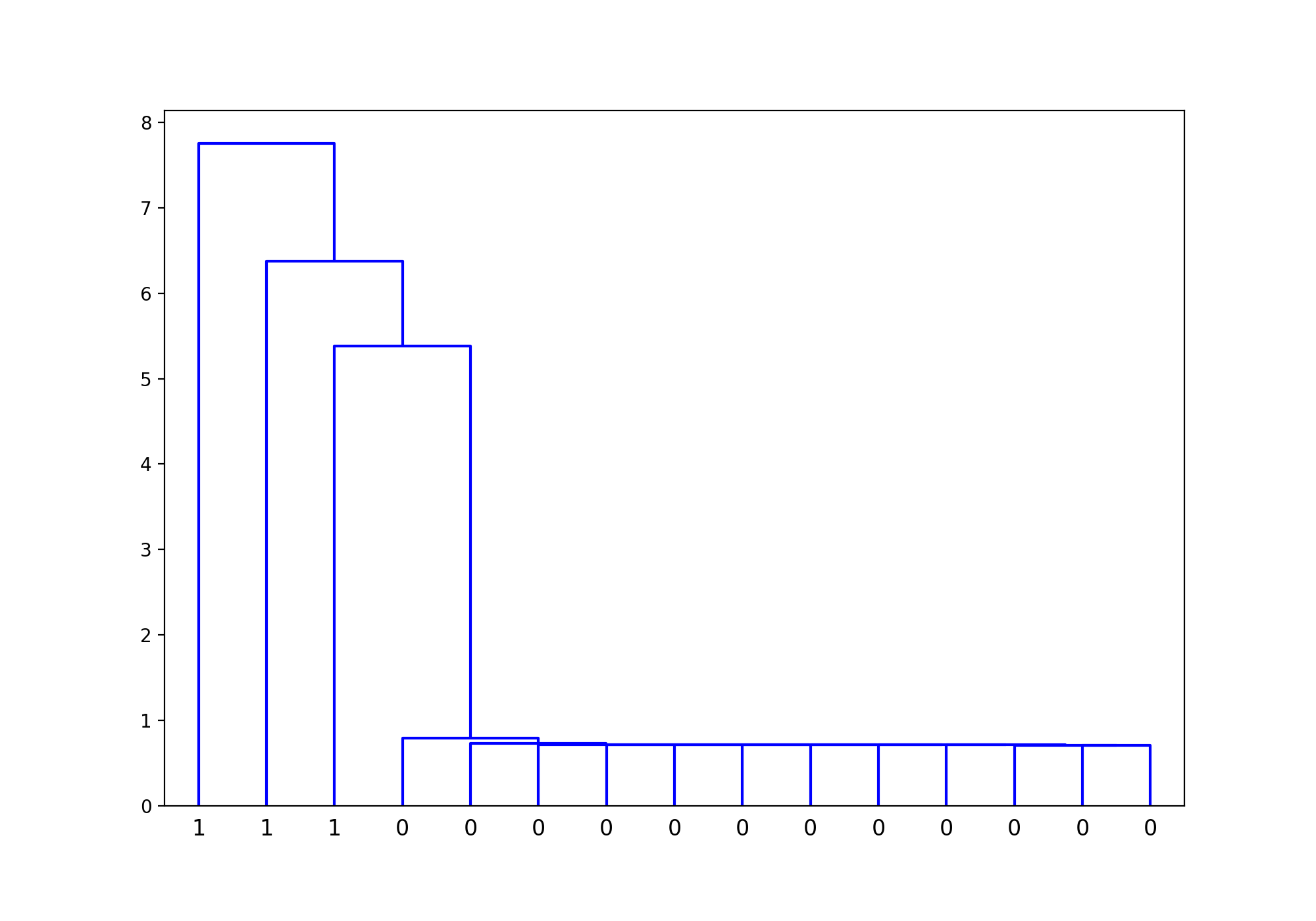}
        \caption{SAX BoP}
        \label{fig:corrupt-nosrkl-run2_hac_sax}
    \end{subfigure}
    
    \begin{subfigure}[b]{0.45\textwidth}
        \centering
        \includegraphics[width=\linewidth,height=140pt]{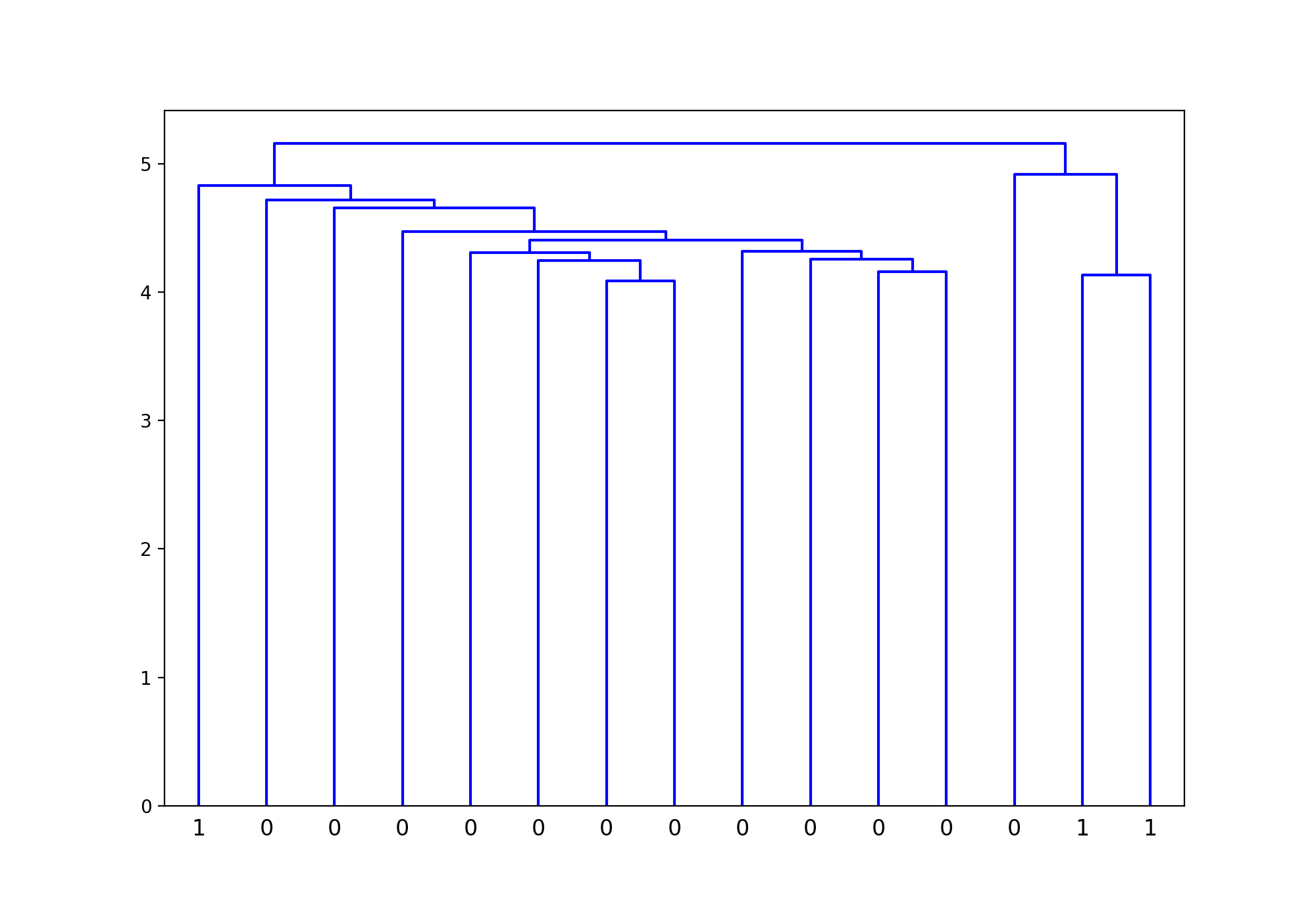}
        \caption{DTW}
        \label{fig:corrupt-nosrkl-run2_hac_dtw}
    \end{subfigure}
    
    \caption{Dendrogram visualizing the HAC clustering of the corrupted gait data found using (a) HRKL, (b) SAX BoP, and (c) DTW. The leaf labels correspond to the grouping labels from \autoref{fig:data_corrupted}.}
    
    
\end{figure}

\autoref{fig:corrupt-nosrkl-run2_hac_gp} illustrates the clustering found using our HRKL, where leaf labels correspond to the grouping labels from \autoref{fig:data_corrupted}. \autoref{fig:corrupt-nosrkl-run2_hac_sax} shows the clustering found using SAX BoP, and \autoref{fig:corrupt-nosrkl-run2_hac_dtw} shows the clustering found using DTW. As previously mentioned, k-Shape is not amenable to a dendrogram representation and requires that the number of clusters be specified beforehand. When initialized with the number of clusters set to two, k-Shape does not recover the same groupings as shown in \autoref{fig:data_corrupted}. Instead, k-Shape achieves a homogeneity score of 0.141, a completeness score of 0.122, and a V-measure score of 0.131.

\autoref{fig:corrupt-nosrkl-run2_hac_gp} shows that the use of our HRKL embedding leads to a clear separation of the corrupt data from the uncorrupted data. \autoref{fig:corrupt-nosrkl-run2_hac_sax} indicates that SAX BoP also clearly separates the corrupted data from the uncorrupted data. However, \autoref{fig:corrupt-nosrkl-run2_hac_dtw} shows that a larger number of clusters, four to be precise, would be required to successfully separate the corrupted data from the uncorrupted data using DTW.

As with the pattern discovery task, since RKL and SRKL do not result in an embedding but instead a single shared kernel, they cannot be used for the anomaly detection task. This again demonstrates the improvement of HRKL over RKL and SRKL, as HRKL can be used for tasks that RKL and SRKL cannot be used for.




\begin{table}[]
\caption{Clustering performance metrics on the corrupted gait data comparing HAC clustering using HRKL, DTW, and SAX BoP, as well as k-Shape, all provided with the correct number of clusters.}
\label{tbl:clus_table_corrupted}
\begin{tabular}{l|llll}
 & Homogeneity & Completeness  & V-Measure  \\ \hline 
 HRKL & \textbf{1.000} & \textbf{1.000} & \textbf{1.000}  \\
 DTW & 0.287 & 0.287 & 0.287 \\
 SAX BoP & 0.235 & 0.480 & 0.316  \\
 k-Shape & 0.141 & 0.122 & 0.131
\end{tabular}
\end{table}

\section{Conclusion} \label{sec:conclusion}

We have extended prior work to create an interpretable kernel embedding for time series which allows for wider flexibility to deal with noisy data that may contain outliers and for the inclusion of sub-population identification as a natural part of an automated statistician. In other words, this embedding allows for heterogeneous relational kernel learning and for automatic hypothesis generation from sets of time series where only subsets of the time series share structure. This embedding can also be used for tasks such as clustering and anomaly detection in sets of time series. We showed the validity of our interpretable kernel embedding on three separate tasks including clustering, pattern discovery, and anomaly detection, where our method was shown to perform well across all tasks and in comparison to other popular time series clustering techniques.

\bibliographystyle{named}
\bibliography{Mendeley,Mendeley-Andre.bib}

\section*{Acknowledgements}

Special thanks to Drew Farris for his support of this work and to Eli N. Weinstein for interesting conversations and insights on connections to model based search.

\end{document}